\documentclass{article}
\usepackage{amsmath,graphicx, url, color, colortbl}
\usepackage{cite, subcaption}
\usepackage[hidelinks]{hyperref}
\hypersetup{
    colorlinks=true,
    linkcolor=black,
    filecolor=black,
    urlcolor=gray,
    citecolor=black,
    breaklinks=true
}
\usepackage{tikz}
\usepackage{pgfplots}
\usepackage[preprint]{spconf}
\pgfplotsset{width=7cm,compat=1.17}
\usepgfplotslibrary{statistics}
\usetikzlibrary{pgfplots.groupplots}

\usepackage{bbm}
\def\1{\mathbbm{1}}
\title{Container Localisation and Mass Estimation with an RGB-D Camera} 
\name{Tommaso Apicella, Giulia Slavic, Edoardo Ragusa, Paolo Gastaldo and Lucio Marcenaro 
}
\address{DITEN, University of Genoa, Italy
}

\begin{document}
\setcounter{page}{0}
\twocolumn[\noindent Copyright 2022 IEEE. Published in ICASSP 2022 - 2022 IEEE International Conference on Acoustics, Speech and Signal Processing (ICASSP), scheduled for 7-13 May 2022 Virtual; 22-27 May 2022 In-Person in Singapore. Personal use of this material is permitted. However, permission to reprint/republish this material for advertising or promotional purposes or for creating new collective works for resale or redistribution to servers or lists, or to reuse any copyrighted component of this work in other works, must be obtained from the IEEE. Contact: Manager, Copyrights and Permissions / IEEE Service Center / 445 Hoes Lane / P.O. Box 1331 / Piscataway, NJ 08855-1331, USA. Telephone: + Intl. 908-562-3966.]
\ninept
\maketitle

\begin{abstract}
In the research area of human-robot interactions, the automatic estimation of the mass of a container manipulated by a person leveraging only visual information is a challenging task. The main challenges consist of occlusions, different filling materials and lighting conditions. The mass of an object constitutes key information for the robot to correctly regulate the force required to grasp the container. We propose a single RGB-D camera-based method to locate a manipulated container and estimate its empty mass i.e., independently of the presence of the content. The method first automatically selects a number of candidate containers based on the distance with the fixed frontal view, then averages the mass predictions of a lightweight model to provide the final estimation. Results on the CORSMAL Containers Manipulation dataset show that the proposed method estimates empty container mass obtaining a score of 71.08\% under different lighting or filling conditions.
\end{abstract}


\section{Introduction}
\label{sec:intro}
Estimating the physical properties of objects is fundamental for the success and safety of human-robot collaboration~\cite{ortenzi2021object}. A robot is a completely autonomous system and is supposed to extract information of the target object in order to assist the human correctly. Robots could be employed, for example, to help people perform housework, lift heavy loads or even bring medicines to the elderly. An incorrect prediction of the object properties could harm the human, e.g., dropping the object or spilling dangerous substances~\cite{pang2021towards}.

In a typical human-to-robot handover scenario, the person exchanges a container with a robot, which takes it from human hand(s)~\cite{ortenzi2021object, rosenberger2021object}. The estimation of container properties such as its width, height and mass represents a crucial stage, since the robot regulates the force to hold the object during the handover and the maneuvering~\cite{pang2021towards}. Moreover, it is not a trivial task since the object could be unknown~\cite{yang2021reactive, rosenberger2021object} or the physical properties of the container could change based on the interaction, e.g., deformation due to the grasp, or different stiffness and filling amounts~\cite{sanchez2020benchmark}.

The container mass can be indirectly retrieved by combining two properties: filling mass and empty container mass. The filling mass can be seen as the result of three contributions: filling type classification, filling level and container capacity estimation~\cite{xompero2020multi}. Recent exisiting solutions use the CORSMAL Containers Manipulation (CCM) dataset~\cite{xompero2020corsmal}, which includes annotated audio-visual recordings of people interacting with containers~\cite{xompero2021multi}.
To perform filling type classification, audio is one of the most used modality, either processing spectrograms~\cite{christmann20202020} or classical audio features~\cite{cho2014properties}. Solutions for filling level estimation can exploit only audio modality~\cite{ishikawa2021audio} or the combination with visual data~\cite{liu2021va2mass,iashin2021top}. Visual clues represent the main modality used to estimate the container capacity. Approaches can consider the task as a regression problem, employing either simple CNNs on single fixed frontal view depth data ~\cite{christmann20202020} or distribution fitting via Gaussian processes using object category as a prior across multiple views~\cite{liu2021va2mass}. Otherwise, the segmented container can be approximated to a primitive shape in 3D, computing capacity as a by-product~\cite{ishikawa2021audio}, or using volume formulas~\cite{iashin2021top}. 

Unlike previous works that focused on estimating the filling mass, we propose an approach to localise the container manipulated by a person and then estimate its empty mass (regardless of the content) from an RGB-D camera with fixed frontal view. Adopting human-to-robot handover as a use case, the strategy consists in automatically selecting a number of containers candidates based on the average distance with respect to the frontal view and then estimating the empty container mass using a lightweight CNN\footnote{Code available at: \url{https://github.com/CORSMAL/Visual}}. A similar CNN was previously devised to regress the container capacity using only depth image crops as input~\cite{christmann20202020}. However, depth images can be noisy, and can contain missing values even after applying basic morphological operations (e.g., closing). Using both geometrical information (distance from depth map and aspect ratio with respect to the original resolution) and the colour information from the RGB frame increases the probability of localising the object and estimating the empty container mass.

\definecolor{tcomp}{RGB}{253,156,111}    %
\definecolor{tmod}{RGB}{229,229,253}
\begin{figure*}[t]
    \centering
    \includegraphics[width=0.8\linewidth]{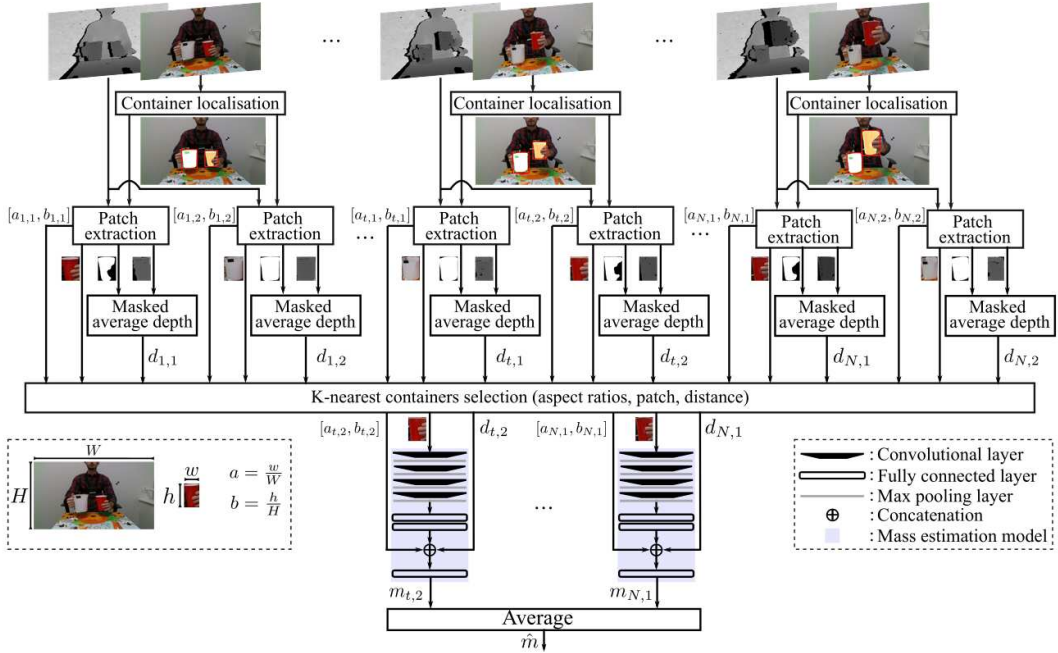}
    \caption{Block diagram of the proposed approach. For each frame, containers are detected, then $K$ nearest patches are selected leveraging the raw depth maps considered in the segmentation mask coordinates. The empty mass of each patch ($m$) is predicted by the model which takes as input the RGB patch and  triplets of values: aspect ratio width ($a$) and height ($b$), and average distance ($d$). The final empty mass estimation ($\hat{m}$) is the average of $K$ mass predictions.  
    }
    \label{fig:procedure}
    \vspace{-10pt}
\end{figure*}

\section{Localisation and mass estimation}
\label{sec:methodology}
The proposed method uses RGB-D data and is divided into three steps (see Fig.~\ref{fig:procedure}).
The first step localises the container every $n$ frames using Mask R-CNN~\cite{he2017mask} pre-trained on COCO~\cite{lin2014microsoft}, selecting only the classes \textit{cup}, \textit{book}, \textit{wine glass} and \textit{bottle}~\cite{iashin2021top}.
Assuming that the manipulated container is the nearest with respect to a fixed frontal view, the method automatically selects the nearest detected containers based on the average distance of the depth values belonging to the container mask (\textit{K-nearest patches selection}).
The last step is the \textit{final prediction} in which the model performs one prediction per each patch and then empty container mass ($\hat{m}$) is computed as the average of the predictions.

\subsection{Mass estimation model}
The model, designed to be lightweight, takes as input RGB patches of the container, the width and height aspect ratios with respect to the original image resolution ($a$ and $b$) and the average distance of the object from the camera ($d$). The rationale behind this last features ($f$) is to preserve some geometrical information of the distance between the container and camera, along with the aspect ratios. Similarly to~\cite{christmann20202020}, the model has four convolutional layers, two Fully Connected (FC) layers, concatenation of the output of the second FC layer with $f$, and one FC layer after the concatenation. Batch normalization and ReLU activations are used after each layer. Max Pooling is used after each convolutional layer. The size of convolutional kernels is $(3,3)$, paddings and strides are $(1,1)$, and channel dimensions are $(32, 64, 64, 128)$. The size of Max pooling kernel is $(2,2)$. The output of the first two FC layers has dimension $64$ and $6$, respectively. Unlike~\cite{christmann20202020} that uses depth patches and their aspect ratios $[a, b]$, our model takes as input RGB patches and the features $f = [a, b, d]$. The number of model parameters is 533,000. 

\subsection{Model training}
\begin{figure}[t]
    \centering
    \includegraphics[width=\columnwidth]{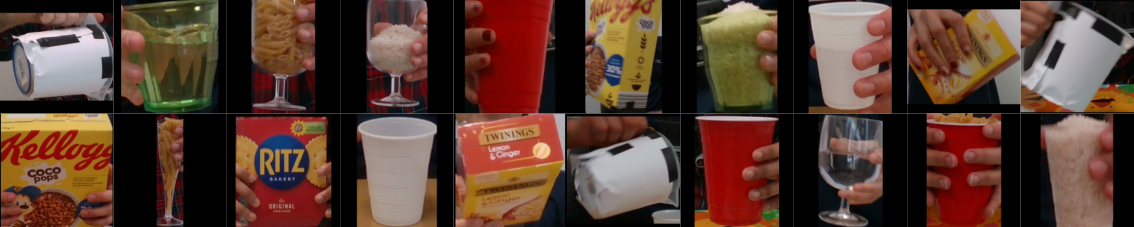}
    \caption{Sample patches of the extracted dataset. Black padding is applied before resizing to keep the same aspect ratio.
    }
    \label{fig:patches}
    \vspace{-10pt}
\end{figure}

The model is trained using the containers patches extracted using the first two steps of the described procedure on the CCM dataset. Mask R-CNN is applied to the entire dataset, similarly to Crop-CCM~\cite{modas2021improving}. The main differences are that we use a single view of the scene, we do not restrict the model solely to cup and glass cases, and we do not perform a manual check of the results. The selection of the containers patches is indeed performed automatically during \textit{$K$-nearest patches selection} phase. Images resolution is $1280 \times 720$, the detection threshold is set to 0.4 and every frame of the video is analysed ($n=1$). In each frame, Mask R-CNN could find in general more than one object e.g. the pitcher used to fill the cup and the cup itself. The maximum number of considered containers candidates ($K$) is set to 5 (note: the model could also detect less patches in a recording). The rationale behind the parameter value choice is to have a trade-off to obtain enough patches to train the model and provide a more robust mass prediction, by averaging a number of candidates, as well as minimizing computational overhead. Our extracted dataset consists of 3,408 patches, some of them are shown in Fig.~\ref{fig:patches}. The proposed automatic retrieval leads to almost 8\% of the patches containing the pitcher used to fill containers, which are not annotated in CCM dataset. The empty mass annotation of each video is applied to each extracted patch. 

The model is trained on the regression task. 
The patches are resized to $112 \times 112$ resolution, using zero padding on the shorter dimension in order to maintain the proportions, and are normalized to [0, 1] range. The following transformations are employed to augment the patches dataset: horizontal flip with probability 50\%, vertical flip with probability 50\%, random rotation between 0 and 180 degrees without cropping the patch, color jitter which consists in randomly changing the brightness into $[0.8, 1.2]$ range, contrast into $[0.8, 1.2]$ range, saturation into $[0.8, 1.2]$ range and hue into $[-0.2, 0.2]$ range. The aspect ratios, average distance and empty mass labels are normalized using the minimum and maximum values retrieved from the training set.
The following setup is common to the experiments: mean square error loss, batch size 32, learning rate 0.0015 with an exponential decay rate of 0.9985 and with decay steps equal to 20; weight decay is set to 0.001.

\section{Experimental setup}
\label{sec:experimentalsetup}
We evaluate our proposed model on the CCM dataset~\cite{xompero2020corsmal} which consists of 1140 audio-visual recordings. During each recording, a person interacts with a container (e.g., filling a cup with rice contained in a pitcher) and then prepares for the handover. Videos differ in conditions such as lighting, person's clothing, hand occlusions. The total number of containers in the dataset is 15: 9 (684 recordings) constitute the training set, the other 6 are evenly split into a public test set (228 recordings) and private test set (228 recordings).

To assess the generalization performances of the method, we leave one instance per category (\textit{box}, \textit{glass}, \textit{cup}) out, creating three folds~\cite{iashin2021top}, see Fig.~\ref{fig:testfolds}. For each testing fold, the training set includes containers belonging to the other two folds, which are split into training and validation sets using 80\% and 20\% as respective percentages of data. The training set is augmented of 3 times using the described transformations. The model is trained using 100 epochs. The validation set is used for model selection and the best model, the one with the lowest mean loss, is kept for testing.
\definecolor{ttrain}{RGB}{0, 0, 0}    %
\definecolor{ttest}{RGB}{166,166,166}
\begin{figure}[t]
    \centering
    \includegraphics[width=\columnwidth]{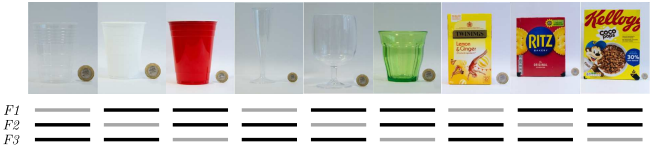}
    \caption{3-fold cross-validation setups ($F1$, $F2$, $F3$) of the CCM training set. Each fold  selects videos from one instance of each container type as test set (\protect\raisebox{2pt}{\protect\tikz \protect\draw[ttest,line width=2] (0,0) -- (0.3,0);}), while videos belonging to the other instances are used as training set (\protect\raisebox{2pt}{\protect\tikz \protect\draw[ttrain,line width=2] (0,0) -- (0.3,0);}).}
    \label{fig:testfolds}
    \vspace{-10pt}
\end{figure}

In addition to this 3-fold cross-validation, we also randomly split the whole training set in training and validation with the 80:20 ratio as before to include all available containers in the training phase. In this case, the method is validated through the CORSMAL Challenge which provides the results for the public test set and privates test set, since the annotations are private. Every patch in the training generates other 4 images using the described transformations. The model is trained for 300 epochs, as the number of training images increases with respect to 3-fold cross-validation. The same rationale of model selection is applied for this experiment. 

As performance measures, we compute for each recording $j$ the relative absolute error between the estimated measure $\hat{m}^{j}$, and the true measure $m^{j}$, as: 
\begin{equation}
    \epsilon(\hat{m}^{j}, m^{j}) = \frac{|\hat{m}^{j} - m^{j}|}{m^{j}}
\end{equation}
The score $s \in [0, 1]$ (where 1 is best), across all $J$ recordings for each measure is\footnote[2]{\url{https://corsmal.eecs.qmul.ac.uk/challenge.html}}:
\begin{equation} \label{eq:score}
    s = \frac{1}{J} \sum_{j=1}^J \mathbbm{1}_{j} \: e^{-\epsilon(\hat{m}^{j},m^{j})}  
\end{equation}
The value of the indicator function $\mathbbm{1}_{j} \in \{0, 1\}$ is 0 only when $\hat{m}$ in recording $j$ is not estimated.

\section{Results}
\label{sec:results}
Fig.~\ref{fig:analysis} (top) analyses the per-class scores (colored based on \textit{cup}, \textit{glass} and \textit{box}) and the \textit{total} score for the three splits (\textit{F1}, \textit{F2}, \textit{F3}) and the validation set (\textit{VAL}). The whole score can be obtained using Eq.~\ref{eq:score} or by multiplying per-class score by the number of instances, summing and then dividing by the total number of instances. Fig.~\ref{fig:analysis} (bottom) provides a complementary analysis through the per-class mean of relative absolute error $\epsilon$. 
Overall, Fig.~\ref{fig:analysis} shows that the model does not generalize to testing containers significantly different from the training ones. The model achieves the highest score and the lowest mean error for the class \textit{box}, probably due to the fact that they feature different colors and have a larger shape with respect to other classes. The mean relative absolute error in the \textit{cup} cases suggests that the empty mass for this class is not properly learned. A possible explanation for the low performance on the first test fold is that the training images contain colored and opaque \textit{cups}, whereas in the test set \textit{cups} are transparent, and the only transparent containers in the used training folds are \textit{glasses}. Other folds statistics point out that the presence of similar containers between training and test sets helps reducing the error and improving the score. The values of mean relative absolute error for \textit{glass} and \textit{box} classes fall in the $[0, 1]$ range, while the \textit{cup} class drops with respect to fold \textit{F1}, yet it remains higher than 1.
The last group of bars in Fig.~\ref{fig:analysis} shows the results for the validation set predictions. The score value underlines that the chosen model is able to learn from recordings having the same containers, yet in different configurations, e.g., lighting conditions or filling. Also in this case, the class that impacts the most on the score is \textit{cup}. 

\pgfplotstableread{example.txt}\scores 
\pgfplotstableread{example2.txt}\error 

\definecolor{tcup}{RGB}{0, 0, 0}    %
\definecolor{tglass}{RGB}{238,113,27}  %
\definecolor{tbox}{RGB}{127,127,127}  %
\definecolor{ttot}{RGB}{229, 194, 36}
\definecolor{tavg}{RGB}{74, 74, 250}
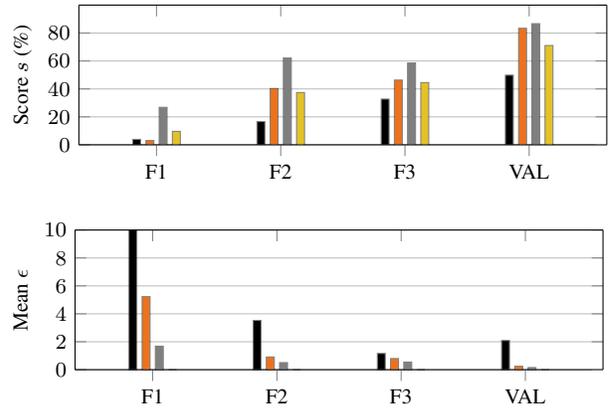
\begin{figure}[t]
    \centering
    \begin{tikzpicture}
    \centering
    \begin{axis}[
    axis x line*=bottom,
    axis y line*=left,
    enlarge x limits=false,
    ybar,
    width=\columnwidth,
    bar width=3pt,
    xmin=0,xmax=17,
    xtick={2.5,6.5,10.5,14.5},
    xticklabels={$\text{F1}$,$\text{F2}$,$\text{F3}$, $\text{VAL}$},
    height=0.4\columnwidth,
    ymin=0,  ymax=100,
    ytick={0,20,40,60,80},
    ylabel={Score $s$ (\%)},
    label style={font=\footnotesize},
    tick label style={font=\footnotesize},
    ymajorgrids=true,
    ]
    \addplot+[ybar, black, fill=tcup, draw opacity=0.5] table[x=CoID,y=cup]{\scores};
    \addplot+[ybar, black, fill=tglass, draw opacity=0.5] table[x=CoID,y=glass]{\scores};
    \addplot+[ybar, black, fill=tbox, draw opacity=0.5] table[x=CoID,y=box]{\scores};
    \addplot+[ybar, black, fill=ttot, draw opacity=0.5] table[x=CoID,y=total]{\scores};
    \end{axis}
    \begin{axis}[ 
        axis x line*=top,
        axis y line*=right,
        width=\columnwidth,
        height=.4\columnwidth,
        xmin=0,xmax=17,
        tick label style={font=\footnotesize, align=center,text width=3cm},
        xtick={2.5,6.5,10.5,14.5},
        xticklabels={},
        typeset ticklabels with strut,
        label style={font=\footnotesize},
        ymin=0,ymax=100,
        yticklabels={},
    ]
    \end{axis}
    \end{tikzpicture}
    \vspace{-10pt}
    \begin{tikzpicture}
    \centering
    \begin{axis}[
    axis x line*=bottom,
    axis y line*=left,
    enlarge x limits=false,
    ybar,
    width=\columnwidth,
    bar width=3pt,
    xmin=0,xmax=17,
    xtick={2.5,6.5,10.5,14.5},
    xticklabels={$\text{F1}$,$\text{F2}$,$\text{F3}$, $\text{VAL}$},
    height=0.4\columnwidth,
    ymin=0,  ymax=10,
    ytick={0,2,4,6,8,10},
    ylabel={Mean $\epsilon$},
    label style={font=\footnotesize},
    tick label style={font=\footnotesize},
    ymajorgrids=true,
    ]
    \addplot+[ybar, black, fill=tcup, draw opacity=0.5] table[x=CoID,y=cup]{\error};
    \addplot+[ybar, black, fill=tglass, draw opacity=0.5] table[x=CoID,y=glass]{\error};
    \addplot+[ybar, black, fill=tbox, draw opacity=0.5] table[x=CoID,y=box]{\error};
    \addplot+[ybar, black, fill=ttot, draw opacity=0.5] table[x=CoID,y=total]{\error};
    \end{axis}
    \begin{axis}[ 
        axis x line*=top,
        axis y line*=right,
        width=\columnwidth,
        height=.4\columnwidth,
        xmin=0,xmax=17,
        tick label style={font=\footnotesize, align=center,text width=3cm},
        xtick={2.5,6.5,10.5,14.5},
        xticklabels={},
        typeset ticklabels with strut,
        label style={font=\footnotesize},
        ymin=0,ymax=10,
        yticklabels={},
    ]
    \end{axis}
    \end{tikzpicture}
   \caption{Analysis per container type of 3-fold cross validation and random cross-validation of our proposed model for container empty mass estimation. Top: testing score $s$ in percentage. Bottom: mean of relative absolute error $\epsilon$. The maximum y-axis value is set to 10 for visualization purpose, the actual value for \textit{cup} in \textit{F1} is 22.952.
   Legend:
    \protect\raisebox{2pt}{\protect\tikz \protect\draw[tcup,line width=2] (0,0) -- (0.3,0);}~\textit{cup},
    \protect\raisebox{2pt}{\protect\tikz \protect\draw[tglass,line width=2] (0,0) -- (0.3,0);}~\textit{glass},
    \protect\raisebox{2pt}{\protect\tikz \protect\draw[tbox,line width=2] (0,0) -- (0.3,0);}~\textit{box}
    \protect\raisebox{2pt}{\protect\tikz \protect\draw[ttot,line width=2] (0,0) -- (0.3,0);}~\textit{total}
   }
    \label{fig:analysis}
    \vspace{-10pt}
\end{figure}

The performance score of CORSMAL challenge methods evaluated on private and public test set are shown in Table~\ref{tab:fscoreschallenge}. Method 1 (\textit{M1}) exploits RGB-D data from the fixed frontal view, extracts object patches using an object detector (YoloV5\footnote[3]{\url{https://github.com/ultralytics/yolov5}}), then predicts the empty container mass using an efficient model (MobileNetV2~\cite{mobileNetV2}) enhanced with attention mechanism and pre-trained on container dimension estimation~\cite{wang2022improving}. The eventual mass prediction is obtained by averaging the mass predictions. Method 2 (\textit{M2}) regresses the empty container mass using a custom CNN which combines: the patch of the container extracted using a formula to select the most visible view (across fontal, left and right views of the scene), the symmetrically restored object mask from left side fixed view, and information about container dimensions (height, width at the bottom, width at the top)~\cite{matsubara2022shared}. The container detection is performed by Localisation and object Dimensions Estimator (LoDE)~\cite{xompero2020multi}. Our model ($M3$) is the one providing \textit{VAL} results of Fig.~\ref{fig:analysis}. As baselines for comparison, we consider also a pseudo-random generator (\textit{M4}) that draws the predictions from a uniform distribution in the interval $[1, 351]$ based on the Mersenne Twister algorithm~\cite{matsumoto1998mersenne}, and average (\textit{M5}) computed on mass labels. In general, some similar features across the methods can be highlighted e.g. the employment of two-stage approach (first detection then mass estimation) to tackle the problem and the use of lightweight models to perform the mass prediction. Our method achieves a higher score with respect to \textit{M2}, \textit{M4} and \textit{M5}. Contrary to \textit{M2}, our method does not use LoDE during the container detection phase and exploits only the fixed frontal view. The generalization properties on private test set show that \textit{M1} performs better than other models, probably due to the employment attention mechanisms. Compared to this solution, our model is not pre-trained on other container properties estimation and features less parameters than MobileNetV2 model.
\begin{table}[t!]
    \caption{Public and private test scores of container mass estimation solutions in percentage.}
    \tabcolsep=0.15cm
    \centering
    \begin{tabular}{ c c c c c c }
        \hline
        \textbf{Set} &  \textbf{M1} & \textbf{M2} & \textbf{M3 (Ours)} & \textbf{M4} & \textbf{M5}   \\\hline
        Public test & 55.25 & 43.61 & 53.14 & 30.59 & 21.88 \\
        Private test & 62.32 & 36.77 & 46.14 & 28.25 & 22.24 \\
        Combination & 58.78 & 40.19 & 49.64 & 29.42 & 22.06 \\\hline
    \end{tabular}
    \label{tab:fscoreschallenge}
     \vspace{-10pt}
\end{table}

\section{Conclusion}
\label{sec:conclusion}
This paper provides a method to analyze a video, extract the container subject to manipulation based on its distance with respect to the fixed frontal view, and estimate its mass regardless of the content using a lightweight model. The low percentage of pitcher patches suggests that the proposed method is able to locate the manipulated container in training recordings. In the experiments, the model learns from similar containers, but poorly generalizes to unseen containers, especially \textit{cups}. As future work, we will investigate multi-task learning to also classify the container type and improve generalization, as well as audio-visual perception to exploit the complementarity of the audio modality. Moreover, the case of camera mounted on the robot will be considered to analyse how partial views of the container due to movement affect model predictions.

\bibliographystyle{IEEEbib}
\bibliography{arxiv}

\end{document}